\documentclass[letterpaper]{article}
\usepackage[preprint]{aaai2027}
\usepackage[hyphens]{url}
\usepackage{graphicx}
\usepackage{natbib}
\usepackage{caption}
\usepackage{amsmath,amssymb,amsthm}
\usepackage{booktabs}
\usepackage{float}
\usepackage{multirow}

\newtheorem{proposition}{Proposition}

\title{Dueling World Models:\\
Advantage-Style Action Channels for Common-Mode Distractor Rejection}

\author{
    Jiazhuo Li\textsuperscript{\rm 1}\equalcontrib\corresponding,
    Yiming Fei\textsuperscript{\rm 2}\equalcontrib,
    Zhiruo Zhou\textsuperscript{\rm 3},
    Heikichi Hayashi\textsuperscript{\rm 4}
}
\affiliations{
    \textsuperscript{\rm 1}University of Michigan, Ann Arbor, MI, USA\\
    \textsuperscript{\rm 2}Zhejiang University, Hangzhou, China\\
    \textsuperscript{\rm 3}Wuhan University of Technology, Wuhan, China\\
    \textsuperscript{\rm 4}Adrasteia Labs\\
    jiazhuo@umich.edu
}

\begin{document}

\maketitle

\begin{abstract}
Latent world models plan by predicting future states from an action, but when a scene contains motion the agent does not control, they quietly go action-blind: predictions for different actions become indistinguishable even as the training loss keeps improving.
Existing remedies suppress this distraction with reconstruction, task reward, or auxiliary objectives, each adding machinery or assumptions.
We show that a minimal alternative suffices, borrowed from the dueling decomposition of value into a state baseline and an action advantage: in latent dynamics, subtracting a prediction's mean effect over actions cancels whatever the actions share---the action-independent variation where distractors live---leaving a clean, controllable channel, with no reward, no reconstruction, and no distractor-specific auxiliary loss.
Because this is only a subtraction at readout time, it applies unchanged to any action-conditioned world model, including frozen pretrained ones.
Across a gridworld, synthetic generators with known factors, distracting continuous control, and natural-pixel Atari, the isolated channel recovers the agent's own effect where entangled predictors fail, with nuisance leak indistinguishable from zero; applied post hoc it surfaces an action channel in off-the-shelf models that their raw readouts miss, and it converts into goal-reaching control in the gridworld.
We prove the cancellation is exact in finite samples for both discrete and sampled action sets, and we state its measured boundary---distractors whose motion tracks the action---together with the remaining limitations in the appendix.
\end{abstract}

\section{Introduction}
\label{sec:intro}

Latent world models \citep{ha2018world,hafner2023mastering} increasingly follow the joint-embedding predictive architecture (JEPA) recipe \citep{lecun2022path}: encode observations, predict future \emph{embeddings} rather than pixels, and plan in the learned latent space \citep{assran2025vjepa2,zhou2025dinowm,maes2026lewm}. Under action-conditioned instances, one monolithic predictor $P(z,a)$ absorbs everything predictable about the next embedding: the consequences of the agent's action, but also the autonomous evolution of whatever else fills the frame.
That distraction degrades learned action and dynamics channels is by now established---latent actions absorb distractor dynamics unless grounded by true action labels \citep{nikulin2025laom} or purged with external segmentation masks \citep{fechner2026segment}.
Existing remedies buy the separation with extra machinery: reconstruction and reward-based factorisation \citep{fu2021tia,wang2022denoised}, separate controllable and uncontrollable branches \citep{pan2022isodream}, bisimulation-style objectives \citep{zhu2023repo}, or an auxiliary inverse-dynamics incentive on a monolithic predictor \citep{ivashkov2026sensorimotor}.
Each adds a loss, a decoder, or an assumption about the distractor process, and none isolates the action channel as a property of the predictor itself. Under action-labeled circumstances, we ask the mechanism question: what structure inside the predictor makes the given actions count?

We first quantify the failure. 
In a $13{\times}13$ FourRooms gridworld we add $n$ rolling binary distractor cells that are action-independent and track \emph{action separation} (AS): the mean pairwise distance between the predictor's outputs for different actions at the same state.
As $n$ grows to $30$, AS collapses from ${\approx}1.28$ to ${\approx}0.002$---for \emph{every} architecture we train, including ours (Fig.~\ref{fig:e1}): the one-step prediction is dominated by the distractor field, and the action's contribution shrinks below numerical relevance.
Probing shows the information is not lost, as the action's effect remains recoverable from the representation. Yet the predictor simply stops routing it anywhere readable.
The failure is silent (embedding-prediction validation loss keeps improving) but not benign: goal-reaching control with the standard predictor degrades sharply, and standard model selection actively prefers the pathology---on Atari Freeway, validation-loss checkpoint selection returned action-collapsed checkpoints in 17 of 36 runs.

Therefore, we propose a fix that restructures the predictor into two streams: a passive stream that predicts how the scene evolves with the action marginalised out, and an action stream that is \emph{centered}, in the sense that the mean prediction over actions is subtracted from it before it is added back.
That subtraction is an algebraic identity rather than a learned tendency: whatever enters the action stream \emph{identically across actions}---common-mode variation, including whatever action-independent distractors contribute---is removed exactly, leaving only what distinguishes one action from another. It survives estimation, staying exact when the mean is taken over sampled actions rather than enumerated (Sec.~\ref{sec:method}).
This is a claim about \emph{per-transition common-mode structure}, not about the distractor process: it needs no process-level exogeneity assumption \citep{efroni2022provably}, no reward, and no reconstruction \citep{fu2021tia,wang2022denoised,pan2022isodream}.
Whether real distractors actually enter as common mode is an empirical question; our results across gridworld and Atari indicate that they overwhelmingly do.
The subtraction is Dueling DQN's identity $Q = V + (A-\bar A)$ \citep{wang2016dueling} transplanted from scalar values to vector latent dynamics, where centering---an identifiability device in the original---acquires a second role: common-mode rejection.
The action-marginal baseline itself appears in \citet{seitzer2021cai} as a post-hoc scalar influence score for exploration, and concurrent work \citep{ivashkov2026sensorimotor} pursues the same distractor-discarding outcome through an inverse-dynamics auxiliary loss on a monolithic predictor.
Our claim is the readout, not the architecture: a vector-valued, common-mode--invariant action channel that centering exposes in any action-conditioned predictor---post hoc on frozen models, or embedded in the parameterization, where its invariance needs no auxiliary objective.
We call the resulting model AD-JEPA (Action-Decomposed JEPA).\footnote{Not to be confused with AD-L-JEPA \citep{zhu2026adljepa}, a self-supervised LiDAR pre-training method for autonomous driving; the acronym collision is coincidental---there ``AD'' abbreviates \emph{autonomous driving}, here \emph{action-decomposed}.}

The same scoping that gives the guarantee defines its limit: distractors whose dynamics correlate with the agent's actions are not common mode and are not cancelled.
We construct such cases and measure the failure---offset probes collapse to chance and nuisance rejection breaks---consistent with the published stress cases, reafferent distractors \citep{hutson2024psp} and agent-like distractors \citep{wang2024ad3}; incentive-based separation concedes the same failure mode \citep{ivashkov2026sensorimotor}.
We report this boundary alongside every positive result.
We also report a negative result on our own ladder: a learned gate stacked on centering is inert (it degenerates to uniform down-scaling rather than sparse selection), so centering alone is the mechanism.

\paragraph{Contributions.}
\begin{itemize}
\item We propose \emph{action-mean centering}, a one-line restructuring of
an action-conditioned latent predictor that makes its action channel
common-mode--invariant, and we analyse it: the cancellation is exact in
finite samples for discrete and sampled action sets alike
(Propositions~\ref{prop:cancel}--\ref{prop:sample}), with a precisely
scoped boundary---action-correlated variation passes through untouched.
\item We give an implementation and validate it across four settings---a
gridworld, synthetic generators with known factors, distracting continuous
control, and natural-pixel Atari---where the centered channel preserves the
agent's own effect with nuisance leak indistinguishable from zero, and
converts into goal-reaching control in the gridworld.
\item Because the mechanism lives in the readout rather than the
architecture, it applies to models we did not train: the same subtraction
surfaces an action channel in frozen RePo and TIA world models whose raw
readouts show none, which suggests such a channel is latent in
action-conditioned predictors generally and merely unrouted
(Sec.~\ref{sec:plugin}).
\end{itemize}

All results use three seeds unless a caption states otherwise (frozen-host
cells: one training run each, readout $\pm$s.d.\ over five probe seeds).

\section{Related Work}
\label{sec:related}

Our contribution intersects three lines of work: advantage-style
decompositions that center predictions on an action marginal, latent-action
and JEPA world models, and factored world models that separate controllable
from exogenous dynamics. That distractors contaminate the action channels of
learned dynamics models is by now an established problem
\citep{nikulin2025laom,fechner2026segment}; we claim a mechanism, not the
problem statement.

Dueling DQN \citep{wang2016dueling} writes $Q = V + (A - \bar{A})$, where
subtracting the action mean is purely an identifiability device for scalar
values. Transplanted to vector-valued latent dynamics, the same subtraction
acquires a second semantics: any variation entering the per-action offsets
\emph{identically across actions} (common-mode) cancels exactly, in the
discrete and the sampled continuous case alike (\S\ref{sec:method}). The
arithmetic---comparing action-conditioned predictions with their mean over
sampled actions---is CAI's \citep{seitzer2021cai}, applied there post hoc as
a scalar causal-influence score for exploration; the Feedback World Model
\citep{an2026feedback} distills counterfactual variation over sampled
actions into per-dimension controllability weights, estimated offline
and applied only during diffusion-policy guidance. Our claim is the promotion of this
baseline from scalar diagnostic to channel: a vector-valued,
common-mode--invariant action readout that applies post hoc to frozen
predictors and, optionally, embeds in the parameterization---where it
exists throughout learning and adds no auxiliary objective.

Action-conditioned JEPA world models---LeWM \citep{maes2026lewm} with
LeJEPA-style regularization \citep{balestriero2025lejepa}, V-JEPA~2-AC
\citep{assran2025vjepa2}, and DINO-WM \citep{zhou2025dinowm}---train or
post-train a single monolithic predictor $P(z,a)$ in which controllable and
exogenous dynamics remain entangled. SMWM \citep{ivashkov2026sensorimotor}
shares our outcome claim---a reward- and reconstruction-free JEPA world
model that discards uncontrollable distractors---but pursues it by
incentive, an inverse-dynamics auxiliary loss on a monolithic transformer;
our separation is structural, an identity of the parameterization rather
than a learned tendency. In the latent-action literature, LAPO
\citep{schmidt2024lapo}, Genie \citep{bruce2024genie}, and DynaMo
\citep{cui2024dynamo} recover actions through inverse/forward-dynamics
bottlenecks but leave the forward predictor undecomposed; LAOM
\citep{nikulin2025laom} shows observation-only latent actions require
action supervision under distractors---precisely our action-labeled regime,
where centering isolates the controllable channel with no further signals.
MaskLAM \citep{fechner2026segment} purges distractors with external
segmentation masks plus reconstruction, presupposing spatial separability.
AC-LAM \citep{wei2026aclam} is the nearest structural relative: an
additivity prior forbids scene-constant offsets in inferred pseudo-actions,
but as a soft constraint with no action-marginal baseline; centering cancels
\emph{any} common-mode component on true actions, as a parameterization.
PLSM \citep{saanum2024plsm} regularizes how the transition depends on the
state; we decompose how it depends on the action---orthogonal axes.

TIA \citep{fu2021tia} and Denoised MDPs \citep{wang2022denoised} factor the
latent state using reward, reconstruction, and independence structure---the
three dependencies centering removes. Iso-Dream and Iso-Dream++
\citep{pan2022isodream,pan2023isodreampp} train separate
controllable/noncontrollable RSSM branches with reconstruction and
inverse-dynamics losses: a learned, incentive-based split where ours is
identity-level cancellation. The Ex-BMDP line
\citep{efroni2022provably,lamb2022acstate,islam2023acro,levine2024multistep}
proves exogenous filtering under process-level exogeneity and finite-state
conditions; we require only per-transition common-mode structure---an
architectural identity per transition, not an asymptotic recovery result. DreamerPro \citep{deng2022dreamerpro}, RePo
\citep{zhu2023repo}, and HRSSM \citep{sun2024hrssm} obtain
reconstruction-free or bisimulation-style robustness without an explicit
controllability channel (the latter two reward-dependent), and InfoPower
\citep{bharadhwaj2022infopower} contrasts action-conditioned against
action-marginal information inside reward-driven training---a variational
cousin of centering. Earlier controllability partitions separate factors via per-factor
policies or adversarial losses
\citep{thomas2017icf,thomas2018disentangling,sawada2018disentangling,kooi2022disentangled};
ours is algebraic. Finally, two works delimit the
boundary of our guarantee: AD3 \citep{wang2024ad3} generatively infers
implicit actions for agent-like distractors, and Policy-Shaped Prediction
\citep{hutson2024psp} introduces the Reafferent DMC benchmark of
action-correlated distractors. Both exemplify the action-modulated regime
that common-mode cancellation explicitly excludes---a failure mode SMWM
concedes as well---and we return to it in \S\ref{sec:limitations}.

\section{Method}
\label{sec:method}

\begin{figure*}[t]
\centering
\includegraphics[width=\textwidth]{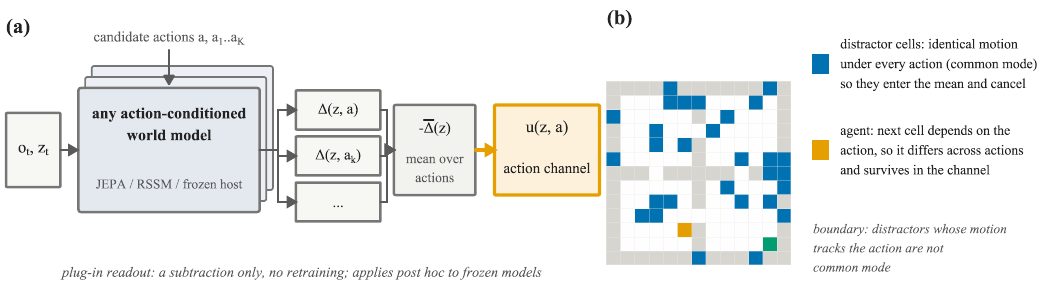}
\caption{\textbf{Centering as a universal plug-in.} (a)~For any
action-conditioned world model---our JEPA variants, an RSSM prior, or a
frozen third-party host---subtracting the mean prediction over candidate
actions yields the effective action channel $u(z,a)$: a readout, applied
post hoc with no retraining. (b)~What ``distraction'' means (FourRooms):
distractor cells move identically under every action (common-mode), so
they enter $\bar\Delta$ and cancel; the agent's next cell depends on the
action and survives in $u$. Action-correlated distractors break the
premise---the stated boundary.}
\label{fig:method}
\end{figure*}

AD-JEPA
learns a latent world model from reward-free, action-labeled transitions
$(o_t, a_t, o_{t+1})$, with no decoder and no reconstruction. The design goal
is not to remove distractor information from the representation, but to
control \emph{where} it lives: the predictor is restructured so that its
action-dependent pathway cannot carry variation that is common across
actions.

\subsection{JEPA Backbone}
\label{sec:method:jepa}

Following the joint-embedding predictive architecture
\citep{lecun2022path}, an online encoder maps observations to latents
$z_t \in \mathbb{R}^d$ and a predictor maps $(z_t, a_t)$ to
$\hat z_{t+1}$; targets come from an EMA target encoder held fixed by a
stop-gradient, as in BYOL \citep{grill2020byol}. Action-conditioned JEPA
world models \citep{zhou2025dinowm, assran2025vjepa2} implement the
predictor as one monolithic network $P(z_t, a_t)$; nothing in that
parameterization separates what the action changes from what would have
happened anyway, and under distractors the two entangle freely. AD-JEPA
is a drop-in restructuring of this predictor.

\subsection{Action-Decomposed Predictor}
\label{sec:method:decomp}

We decompose the predictor into a passive head and a centered
action-offset head:
\begin{equation}
\label{eq:decomp}
\hat z_{t+1} \;=\; B(z_t) \;+\;
  \underbrace{\bigl[\Delta(z_t, a_t) - \bar\Delta(z_t)\bigr]}_{u(z_t,\,a_t)},
\end{equation}
where $\bar\Delta(z) = \mathbb{E}_{a'\sim\nu}\!\left[\Delta(z, a')\right]$ is
the mean offset under a proposal $\nu$ over actions. The passive head $B$
predicts how the scene evolves with the action marginalized out; the
\emph{effective action channel} $u(z,a) = \Delta(z,a) - \bar\Delta(z)$
carries only the action-contrastive part of the dynamics. (The \emph{gated} ablation of Sec.~\ref{sec:protocol} additionally scales the channel elementwise by
a learned $C(z)\in(0,1)^d$; Appendix~\ref{app:gateablation} shows this gate degenerates
to uniform down-scaling, and we report it as a negative result.)

The construction transplants the dueling decomposition
$Q = V + (A - \operatorname{mean}_a A)$ of \citet{wang2016dueling} from
scalar values to vector latent dynamics, where the subtraction acquires a
second semantics. As in dueling Q-learning it is an identifiability
device---for any $c(z)$, replacing $\Delta \mapsto \Delta + c$ leaves
$u$ unchanged, so the channel is well defined though $\Delta$ alone is
not---\emph{and} it performs exact common-mode rejection
(Proposition~\ref{prop:cancel}). Without it, the same degeneracy lets
gradient descent park arbitrary state- and distractor-dependent variation
in the offsets, and empirically it does (Secs.~\ref{sec:e4},
\ref{sec:e6}).

\paragraph{Estimating the action mean.}
For discrete action sets $\mathcal{A}$ we take $\nu$ uniform and enumerate:
$\bar\Delta(z) = \frac{1}{|\mathcal{A}|}\sum_{a'\in\mathcal{A}}
\Delta(z,a')$, computed exactly at every step. For continuous actions, enumeration
is unavailable and $\bar\Delta$ is replaced by a Monte-Carlo mean over
$K{=}16$ candidate actions: the executed action, with the remaining $K{-}1$
split evenly between actions resampled from the replay buffer and Gaussian
perturbations of the executed action (standard deviation $0.3$ of the action
half-range, clipped to the action bounds). Proposition~\ref{prop:sample} (Sec.~\ref{sec:method:cancel}) shows this
estimate leaves the cancellation untouched---exact at any finite $K$---and
confines the $O(1/\sqrt{K})$, proposal-dependent sampling error to the
\emph{centering point}, an action-independent shift; only under discrete
enumeration is the centering point itself exact. The conditioning of
$\bar\Delta$ on $z$ is load-bearing: on frozen hosts, subtracting the
\emph{global} mean offset instead leaves the channel unrecovered
(Appendix~\ref{app:hoststeer})---the common mode being cancelled is
state-conditional. The arithmetic of comparing
action-conditioned predictions against their sampled mean is CAI's
\citep{seitzer2021cai}, applied there as a post-hoc scalar influence
score; AD-JEPA promotes it to a trained, vector-valued channel centered at
every forward pass.

\subsection{Training Objective}
\label{sec:method:losses}

The model is trained with
\begin{equation}
\label{eq:loss}
\begin{split}
\mathcal{L} = {}& \mathcal{L}_{\mathrm{pred}}
  + \lambda_{\mathrm{act}}\,\mathcal{L}_{\mathrm{act}}
  + \lambda_{\mathrm{off}}\,\mathcal{L}_{\mathrm{off}} \\
  &{}+ \lambda_{\mathrm{reg}}\,\mathcal{L}_{\mathrm{reg}}
  \;(+\,\lambda_{\mathrm{gate}}\,\mathcal{L}_{\mathrm{gate}}),
\end{split}
\end{equation}
with the gate term active only in the gated variant.
$\mathcal{L}_{\mathrm{pred}} = 2 - 2\cos\!\bigl(\hat z_{t+1},
\operatorname{sg}[\tilde z_{t+1}]\bigr)$ is the scale-free JEPA prediction
loss against the stop-gradient EMA target.
$\mathcal{L}_{\mathrm{act}}$ is an InfoNCE term \citep{oord2018cpc}
over the $K$ counterfactual-action predictions of Eq.~\eqref{eq:decomp}:
a cross-entropy over cosine similarities to the target requires the
executed action's prediction to be closest---penalizing action-insensitive
predictors.
$\mathcal{L}_{\mathrm{off}}$ is a compactness penalty, the squared norm of
the executed action's effective offset, discouraging the channel from
absorbing state persistence ($\lambda_{\mathrm{off}} = 10^{-4}$).
$\mathcal{L}_{\mathrm{reg}}$ is a VICReg-style anti-collapse regularizer on
the online embeddings---batch-mean centering, a variance hinge, and a
covariance penalty \citep{bardes2022vicreg}---with
$\lambda_{\mathrm{reg}} = 0.1$. For the gated variant,
$\mathcal{L}_{\mathrm{gate}}$ is an L1 penalty on the gate values
($\lambda_{\mathrm{gate}} = 10^{-5}$), held at zero for the first 20\% of
training and then ramped in. All variants train on fixed offline datasets
under identical schedules; data collection, architectures, remaining
hyperparameters, and the action-separation--gated checkpoint-selection
protocol are in Appendix~\ref{app:impl}.

\subsection{Common-Mode Cancellation}
\label{sec:method:cancel}

The property we claim for Eq.~\eqref{eq:decomp} is an identity of the
parameterization, not an incentive supplied by a loss.

\begin{proposition}[Common-mode cancellation]
\label{prop:cancel}
Fix $z$ and suppose the offset head decomposes as
$\Delta(z,a) = g(z,a) + h(z)$ for some $g$ and some action-independent
$h$. Then, with the exact action mean
$\bar\Delta(z) = \frac{1}{|\mathcal{A}|}\sum_{a'\in\mathcal{A}}
\Delta(z,a')$, for every action $a$
\begin{equation*}
\Delta(z,a) - \bar\Delta(z)
  \;=\; g(z,a) - \tfrac{1}{|\mathcal{A}|}\textstyle\sum_{a'} g(z,a'):
\end{equation*}
the common-mode component $h$ cancels exactly, for every value of the
parameters and regardless of what $h$ encodes.
\end{proposition}

\begin{proof}
$h(z)$ appears once in $\Delta(z,a)$ and once in every term of the mean, so
it is subtracted exactly.
\end{proof}

The same holds for the sampled estimate of the continuous regime:

\begin{proposition}[Sampled centering]
\label{prop:sample}
Let $\bar\Delta_K(z) = \frac{1}{K}\sum_{k=1}^{K}\Delta(z, a_k)$ for
candidate actions $\{a_k\}_{k=1}^{K}$ drawn from a proposal $\nu$, and
suppose $\Delta(z,a) = g(z,a) + h(z)$ as above. Then, for every $K$ and
every draw: (i) $\Delta(z,a) - \bar\Delta_K(z) = g(z,a) -
\frac{1}{K}\sum_{k} g(z,a_k)$---the common-mode component $h$ cancels
exactly; and (ii) the sampling error $\bar\Delta_K(z) -
\mathbb{E}_{a'\sim\nu}[\Delta(z,a')]$ does not depend on the query
action, is $O(1/\sqrt{K})$ under bounded second moments of $g(z,\cdot)$,
and cancels in any action-differential comparison $u(z,a) - u(z,a')$.
(Proof and remarks on self-inclusion and proposal bias:
Appendix~\ref{app:proofs}.)
\end{proposition}

Three remarks scope this statement. \emph{First}, the identity is
per-transition and architectural---it holds at initialization, at every
step, and at convergence, with no process-level exogeneity assumption
\citep{efroni2022provably}. Because $\hat z_{t+1} = B + u$ must still
match the target, common-mode variation is routed into $B$: the model is
not blind to distractors---$z$ may represent them and $B$ must track
them---but $u$ has nowhere to hold them. Rejection is routing, not
blindness (Sec.~\ref{sec:e6}). \emph{Second}, the guarantee is exactly
as strong as the common-mode premise: action-correlated (reafferent)
variation enters $g$ and passes through untouched
(Sec.~\ref{sec:limitations}); whether real distractors enter as
common-mode is empirical---Sec.~\ref{sec:e4} tests it with known
factors, and Secs.~\ref{sec:e1} and~\ref{sec:e6} show gridworld and
Atari distractors do behave as common-mode in practice. \emph{Third},
the separation is a property of the function class, not the training
signal: no auxiliary objective rewards discarding uncontrollable
variation, in contrast to inverse-dynamics incentives on a monolithic
predictor \citep{ivashkov2026sensorimotor}.

\section{Experiments}
\label{sec:experiments}

We ask three questions in turn. Does the centered channel survive
distraction where entangled predictors lose it, across environments of
different character? What exactly does it contain, when the ground-truth
factors are known? And is the effect a property of the readout, so that it
transfers to models we did not build? Every configuration uses three seeds;
tables report means, with s.e.m.\ where error bars are shown.

\subsection{Setup and Evaluation Protocol}
\label{sec:exp-setup}\label{sec:protocol}

\paragraph{Variant ladder.}
All variants share the encoder, EMA targets, and losses of
Sec.~\ref{sec:method} and differ only in the predictor: \emph{standard}
$\hat z' = P(z,a)$; \emph{residual} $z + \Delta(z,a)$; \emph{noncentered}
$B(z) + \Delta(z,a)$; \emph{centered} (ours)
$B(z) + [\Delta(z,a) - \bar\Delta(z)]$; and \emph{gated}, which adds a
learned channel gate $C(z)$ and proves inert
(Appendix~\ref{app:gateablation}), so centering is the operative mechanism
throughout. The noncentered$\,\to\,$centered rung isolates the contribution
under test. In the discrete-action environments $\bar\Delta$ is the exact
uniform mean over all actions.

\paragraph{Probing protocol.}
We measure \emph{where} information lives, not just whether it is present:
for the encoder representation $z$, the raw predicted offset, and the
\emph{effective offset} $[\Delta - \bar\Delta]$, we fit ridge probes to
ground-truth factors (5-fold CV, held-out $R^2$). Comparing features
distinguishes \emph{routing} from \emph{blindness}: a factor can be
decodable from $z$ yet absent from the effective offset. Action separation
(AS) is the mean pairwise distance among per-action predictions from a
common state; AS ${\to}\,0$ means the predictor's \emph{output} ignores the
action. Because validation loss alone is an unsafe selection signal
here---on Freeway, cosine-loss selection picked action-collapsed
checkpoints in 17 of 36 runs---we gate checkpoint selection on AS with a
regime-aware threshold. Environments, budgets, distractor knobs and all
per-experiment hyperparameters are in Appendix~\ref{app:impl}.

\subsection{The Centered Channel Survives Distraction}
\label{sec:survives}\label{sec:e1}\label{sec:e2}\label{sec:e6}\label{sec:e3dmc}

Three environments of different character make the same point: as
distraction grows, the centered offset keeps decoding the agent's own
effect while the uncentered offset is destroyed.

\begin{figure*}[t]
  \centering
  \includegraphics[width=\textwidth]{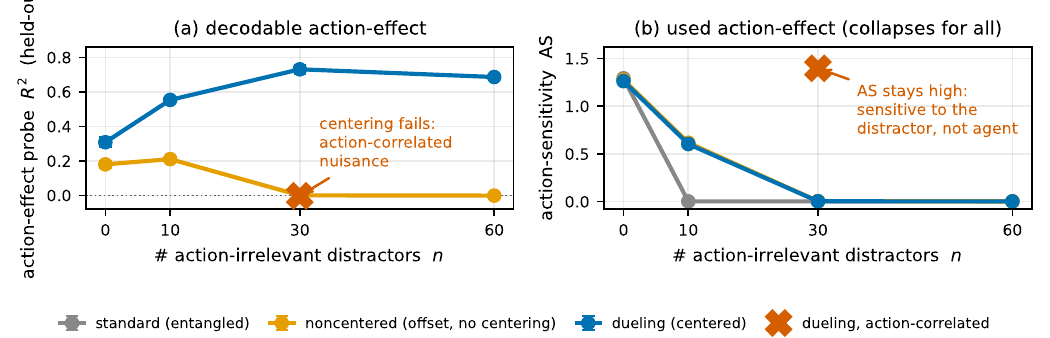}
  \caption{\textbf{Gridworld sweep} ($n$ rolling distractor cells, 3 seeds).
  \emph{Left:} the centered offset keeps decoding the agent's effect while
  the noncentered offset collapses. \emph{Right:} output-level action
  separation collapses for \emph{every} variant---the channel is preserved
  but unused. Legends use the code name \emph{dueling} for centered.}
  \label{fig:e1}
\end{figure*}

\paragraph{Constructed distractors (gridworld).}
In a $13{\times}13$ FourRooms gridworld we add $n$ rolling binary
distractor cells whose dynamics do not depend on the action, so per
transition their contribution to the offsets is common-mode. Across
$n = 0/10/30/60$ the centered offset probes the agent factor at
$0.31/0.55/\mathbf{0.73}/\mathbf{0.69}$ while the noncentered offset falls
to $0.18/0.21/0.00/{-0.00}$ (Table~\ref{tab:e1},
Appendix~\ref{app:floats}); in the clean environment the gap is modest, so
the benefit is specifically common-mode rejection rather than a generic
representation improvement. The channel is preserved but \emph{unused}: AS
collapses from ${\approx}1.28$ to ${\approx}0.002$ for \emph{every}
variant by $n{\ge}30$ (Fig.~\ref{fig:e1}), because once distractor variance
dominates the target the one-step objective is nearly indifferent to the
action.

\paragraph{Pixel distraction (continuous control).}
On three DMC tasks \citep{tassa2018deepmind} we composite $16$
action-independent moving occluders onto the observation---a pixel-level
analogue of Distracting-Control distraction \citep{stone2021distracting}
that leaves the physics-state probe target clean
(Fig.~\ref{fig:dmcframes})---with $\bar\Delta$ a $K{=}16$ Monte-Carlo
mean. Per-seed sign consistency is full (9/9): under independent
distraction the
centered channel probes at $0.25$--$0.53$ against $0.00$--$0.13$
noncentered (Table~\ref{tab:e3dmc}, Appendix~\ref{app:floats}). As retention, centering keeps
95--106\% of its own distraction-free accuracy where the noncentered
ablation keeps 32\% and 2.6\%. Pre-registered criteria and their verbatim
outcomes are in Appendix~\ref{app:e3gate}.

\paragraph{Natural distractors (Atari).}
On \emph{Freeway} \citep{bellemare2013ale} the distraction is generated by
the environment itself---ten lanes of streaming traffic---with RAM probe
ground truth \citep{anand2019atariari} and frameskip as the
distractor-strength knob. The centered offset decodes traffic change and
traffic state at $R^2 \in [-0.001, 0.000]$ at every frameskip, an exactly
clean channel on natural pixels, while keeping the agent's displacement
decodable \emph{above} the action-identity ceiling
($0.877/0.838/0.795$ vs.\ $0.810/0.757/0.683$; Table~\ref{tab:e6_freeway}, Appendix~\ref{app:floats}).
Rejection is routing, not blindness: traffic remains decodable from the
full latent at $R^2 \approx 0.87$ for every variant. The noncentered
ablation is not merely leaky---the leak crowds out the signal, dropping the
agent's own displacement to $0.149$--$0.485$, below even the ceiling.

\paragraph{The channel converts to behaviour where the model is accurate.}
Goal-reaching model-predictive control in latent space, on the same
gridworld sweep, holds at $0.92$--$0.93$ for the centered variants across
$n{=}10$--$60$ where the standard predictor falls to $0.57{\pm}0.03$
(Table~\ref{tab:e2}, Appendix~\ref{app:floats}). Control therefore survives
the AS collapse above: the planner adds no robustness mechanism, it
converts the channel that centering preserves. A pre-registered attempt to
repeat this at DMC scale returns a null, reported in full in
Sec.~\ref{sec:limitations}.

\paragraph{The boundary is shared, and predicted.}
Making the distractors move \emph{with} the action breaks the common-mode
premise, and every readout fails together: the gridworld offset probe drops
to $-0.00$ and control collapses to $0.55$--$0.57$ for all variants at
exactly the configuration where the probe reads chance, and the
continuous-action gate breaks rejection likewise. The representation
boundary predicts the control boundary, and both fall where the analysis
says they must.

\subsection{What the Channel Contains: Recovery without Leakage}
\label{sec:identify}\label{sec:e4}\label{sec:e4n}\label{sec:stage0}

The environments above show the channel survives; they cannot show
\emph{what} it keeps, because the true factors are not separately
observable. We therefore build generators whose factors are known by
construction: a controllable displacement $\Delta_{\mathrm{c}}$ that
depends on the action, a nuisance $\Delta_{\mathrm{w}}$ that does not, and
an encoder that sees only a mixed observation. The nuisance is
action-independent \emph{per transition} (common-mode), not exogenous at
the process level \citep{efroni2022provably}. Two probes score the channel:
held-out $R^2$ against $\Delta_{\mathrm{c}}$ (\emph{recover}, $\uparrow$)
and against $\Delta_{\mathrm{w}}$ (\emph{reject}, $\downarrow$), over 120
configurations spanning both generators and factor dimensionalities.

Only the centered variant does both, at $R^2_{\mathrm{c}} = 0.87$--$0.93$
with $R^2_{\mathrm{w}} \approx -0.00$ across the whole grid
(Table~\ref{tab:e4}, Fig.~\ref{fig:e4}, Appendix~\ref{app:floats}). The
noncentered ablation is the diagnostic case: it recovers nearly as well but
leaks heavily ($0.74$--$0.97$), so predictive quality and channel purity
are different properties. The standard predictor loses the controllable
factor outright once the nuisance dominates. Residual structure buys
recovery; subtracting $\bar\Delta$ buys rejection.

\paragraph{Purity is not an artifact of unpredictability.}
A skeptic can attribute a low nuisance-$R^2$ to unlearnability---no model
predicts an unpredictable nuisance---so we sweep the nuisance from fully
predictable to nearly stochastic and measure leak two ways: against the
nuisance displacement and against the nuisance \emph{state}. The confound
is real (Table~\ref{tab:e4n}): the noncentered $\Delta$-leak decays
$0.947 \to 0.022$ as the nuisance becomes stochastic---read naively,
stochasticity ``purifies'' the model---while its state-leak persists at
$0.80$, because the channel still carries where the nuisance \emph{is}.
Centered is flat at every predictability level, as expected of a
per-transition cancellation rather than an incentive that depends on what
is learnable. Hence the rule we apply throughout: under stochastic
nuisance, measure leak against the nuisance state.

\paragraph{The sampled estimator preserves the picture.}
Repeating the same generator with continuous actions and the $K{=}16$
Monte-Carlo mean of Sec.~\ref{sec:method}, centered recovers at $0.87$ and
rejects at $-0.00$; noncentered recovers while leaking ($0.96$); standard
does neither (Table~\ref{tab:stage0}, Appendix~\ref{app:floats}). The
failure that does register is the intended one---an action-correlated
nuisance breaks rejection to $+0.19$---so the binding assumption is
common-mode structure, not the quality of the estimate.

\subsection{Centering as a Plug-in: Frozen RePo and TIA Hosts}
\label{sec:plugin}

\begin{figure}[t]
\centering
\includegraphics[width=\columnwidth]{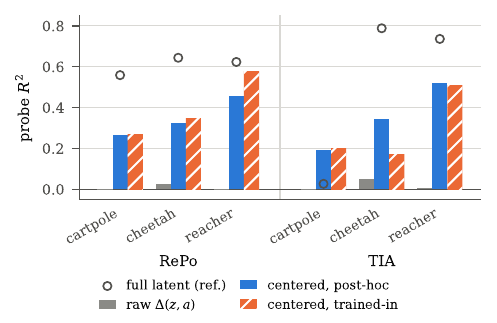}
\caption{\textbf{Centering as an inference-only plug-in} (frozen
RePo/TIA, 500k steps, DAVIS video). Raw action-delta probes ${\approx}0$;
the same model's centered readout recovers $R^2$ $0.19$--$0.52$; trained-in
(hatched) matches or exceeds post hoc on five of six cells. Open markers:
full-latent reference. Per-cell numbers: Table~\ref{tab:plugin}.}
\label{fig:pluginbar}
\end{figure}

The mechanism is a property of the readout, not of our architecture: any
world model exposing an action-conditioned latent transition admits the
same subtraction. We test this on RePo \citep{zhu2023repo} and TIA
\citep{fu2021tia}, each trained 500k steps on the three DMC tasks under
DAVIS video \citep{ponttuset2017davis}---their own regime, not
ours---comparing three readouts of the
\emph{same} frozen model under one held-out ridge protocol: the full
latent, the raw action-delta $\Delta(z,a)$, and the post-hoc centered
channel $\Delta(z,a) - \bar\Delta(z)$ ($K{=}16$ uniform-action
mean)---inference-only, no retraining, no training data.

The raw action-delta reads ${\approx}0$ on every cell ($-0.002$ to
$+0.052$) while the centered readout of the same frozen predictor reaches
$0.19$--$0.52$ (Fig.~\ref{fig:pluginbar}; per-cell numbers in
Table~\ref{tab:plugin})---including on TIA-cartpole, whose
full latent has collapsed ($0.03$) yet still yields a $0.19$ centered
channel. Training the centering in (both hosts, same budget and regime)
passes the same bar on all six cells; against the post-hoc readout it
matches or exceeds on RePo ($0.268$/$0.349$/$0.577$ vs.\
$0.264$/$0.323$/$0.457$) and is comparable on TIA for cartpole and reacher
($0.198$/$0.506$ vs.\ $0.190$/$0.519$) but weaker on TIA-cheetah ($0.168$
vs.\ $0.345$): the mechanism's benefit is fully available at readout time
on a frozen model, and training it in is neither required nor uniformly
better. In the trained-in models, linearly decodable action-effect
information concentrates in the offset channel---the full-latent probe
falls to between $-0.04$ and $0.13$ across all six cells (frozen
baselines: $0.03$--$0.79$); we return to what this concentration implies
in Sec.~\ref{sec:limitations}.

\paragraph{The matched baseline: post-hoc centering of our own standard
predictor.}
If the mechanism is a readout it must work without our architecture:
for any monolithic predictor, $u_P(z,a) = P(z,a) -
\operatorname{mean}_{a'}P(z,a')$ can be formed at inference time. On
FourRooms the centered readout of the plain \emph{standard} predictor
probes at $0.23/0.66/0.70/0.59$ across $n{=}0/10/30/60$ vs.\
$0.31/0.54/0.71/0.66$ trained-in (differences $\le 0.12$ either way), and
resurrects the trained noncentered variant's dead channel
($0.00 \to 0.69$ at $n{=}30$). On DMC the parity is exact to two
decimals ($0.533/0.328/0.186$ vs.\ $0.528/0.323/0.192$,
reacher/cheetah/cartpole), and within one model the contrast is stark:
the standard predictor's raw one-step-change probe reads $-9.2$ (cheetah,
video distraction) while the \emph{same} frozen model's centered readout
reads $+0.33$. The action-correlated boundary is readout-independent
(every post-hoc channel collapses on d30ac). The channel is a property of
action-conditioned prediction itself; the trained decomposition is one
convenient instantiation---the one that carries the control result---not a requirement.
For existing models we recommend the inference-time form: retrofitting
the decomposition into a host's recurrence degrades its control (returns
fall on five of six RePo/TIA cells), while the post-hoc readout leaves
behavior untouched by construction.

\paragraph{Perturbation test with controls.}
Probes are correlational, so we also test the channel's \emph{direction}:
push the host's state along the post-hoc centered offset and decode the
induced physics displacement (first-order pushes; the host is not rolled
forward---direction, not control; Appendix~\ref{app:hoststeer}). On the five
hosts whose instrument validates, the decoded displacement aligns with the
action's counterfactual-centered effect in four (cosine $+0.09$ to
$+0.41$); an equal-norm random push scores ${\approx}0$, permuting the
action pairing collapses the alignment, and subtracting a \emph{global}
mean instead of the state-conditional marginal recovers nothing. The
fifth cell is an instructive null---that host keeps physics in its
deterministic belief---and across hosts the alignment tracks the pushed
block's physics content ($\rho{=}0.90$, $n{=}5$).

\section{Discussion}
\label{sec:discussion}

\subsection{Limitations}
\label{sec:limitations}

Centering cancels exactly what enters the action offsets identically
across actions; it needs no process-level exogeneity assumption
\citep[cf.][]{efroni2022provably}, but by the same token it offers no
guarantee once a distractor's dynamics are modulated by the agent's
action. Our stress tests locate this boundary: in the action-correlated
FourRooms configuration the centered-offset probe collapses to chance and
control degrades for every variant alike, and rejection breaks in the
continuous gate (Secs.~\ref{sec:e1}, \ref{sec:e2}, \ref{sec:stage0})---a
failure mode shared by the published reafferent and agent-like stress
cases \citep{hutson2024psp,wang2024ad3} and conceded by incentive-based
separation \citep{ivashkov2026sensorimotor}; what distinguishes centering
is that the boundary is legible in the algebra. Nor does the identity itself guarantee distractor removal: it removes
what the \emph{learned offsets} represent identically across actions
(a zero-mean interaction $s_{\mathrm{dist}}(z)\phi(a)$ would
survive)---so the channel is common-mode--rejecting by construction,
distractor-rejecting only as measured (Secs.~\ref{sec:e4},
\ref{sec:e6}). Finally, the preserved channel does not by itself restore
control at scale: a multi-step planner converts it into behavior in
FourRooms, but a pre-registered DMC-scale attempt missed its gate across
a $4\times$ training and $32\times$ planning ladder---a null whose
signature (transient-correct steering, late rollout drift) points to
model accuracy rather than a channel defect, though it does not exclude a
composition failure at the control horizon (Appendix~\ref{app:discussion});
a frozen-model perturbation corroborates the channel side
(Appendix~\ref{app:steer}). The control claim is therefore sufficiency,
never superiority at scale; the payoff we defend is
representation-level---the post-hoc channel on frozen predictors
(Sec.~\ref{sec:plugin}). Centering marginalizes only the ego action, so
entities that react to the agent inherit the reafferent boundary.

\subsection{Outlook}
\label{sec:future}

The decomposition is not tied to one-step prediction: a successor-feature
head \citep{dayan1993successor,barreto2017successor} written in centered
form, $\Psi(z,a) = \Psi_B(z) + [\Psi_\Delta(z,a) -
\mathbb{E}_{a'}\Psi_\Delta(z,a')]$, extends common-mode rejection to
accumulated features and, with rewards linear in features, reproduces the
dueling value decomposition---the analogy closes. A common-mode--invariant
$\Psi_\Delta$ is a natural substrate for generalized policy improvement;
validating this, and scaling the representation-to-control conversion, is
future work.

\section{Conclusion}
\label{sec:conclusion}

Action-mean centering restructures an action-conditioned latent
predictor as $\hat z' = B(z) + [\Delta(z,a) - \bar\Delta(z)]$: any
variation entering the action offsets identically across actions
cancels---exactly, for discrete and sampled action sets alike---with no
reward, no reconstruction, and no distractor-specific supervision. Across synthetic, gridworld, distracting-DMC, and Atari testbeds, the
centered channel keeps the agent's own effect decodable where entangled
predictors go action-blind, with nuisance leak indistinguishable from
zero. The mechanism is a property of
the readout, not the architecture: applied post hoc, the same subtraction
recovers an action-effect channel from a plain entangled predictor and
from frozen RePo and TIA models---no retraining. Control gains come from
the passive/offset decomposition rather than centering and do not yet
extend beyond the gridworld; the channel's value is representational, and
its boundary---action-correlated distractors---is explicit in its
algebra. A learned gate proved inert and is reported as a negative
result. We hope centering becomes a default readout for
action-conditioned world models.

\bibliography{refs}

\clearpage   
\appendix

\section{Implementation and Training Details}
\label{app:impl}

\paragraph{Shared setup.}
All variants share the encoder, an EMA target encoder ($\tau{=}0.996$,
buffers copied), the optimizer (AdamW, base learning rate $3\times10^{-4}$,
weight decay $10^{-6}$, cosine-annealed over training with $T_{\max}$ equal
to the step budget, gradient-norm clip $10$), and the loss weights
($\lambda_{\mathrm{act}}{=}0.1$ with softmax temperature $0.1$,
$\lambda_{\mathrm{off}}{=}10^{-4}$, $\lambda_{\mathrm{gate}}{=}10^{-5}$ with a
warm-up over the first $20\%$ of steps, $\lambda_{\mathrm{reg}}{=}0.1$). Only
the predictor head differs across the ladder (Sec.~\ref{sec:method:decomp}).

\paragraph{Architecture.}
The MLP encoder flattens the observation and applies
$\mathrm{Linear}{\to}\mathrm{SiLU}{\to}\mathrm{Linear}{\to}\mathrm{SiLU}
{\to}\mathrm{Linear}$ (two hidden layers of width $256$) to dimension $d$,
followed by \texttt{LayerNorm} \citep{ba2016layernorm}. The CNN encoder applies three
$3{\times}3$ convolutions ($c{\to}32{\to}64{\to}64$ channels, strides
$1,2,2$, padding $1$, SiLU), flattens, and maps to $d$ through a
$256$-wide hidden layer and \texttt{LayerNorm} (a $64{\times}64$ input
reduces to $16{\times}16{\times}64$ before the head). Every prediction head
is a two-hidden-layer, width-$256$ SiLU MLP: the passive head $B{:}~d{\to}d$;
the discrete offset head $d{\to}|\mathcal{A}|{\cdot}d$ (all offsets in one
pass, so $\bar\Delta$ is exact and free); the continuous offset head
$(d{+}m){\to}d$; and the entangled baseline $P$, which embeds the action to
$32$ dimensions (learned embedding for discrete, linear map for continuous)
before an MLP $(d{+}32){\to}d$. The gate $C$ is a one-hidden-layer MLP to a
sigmoid (or a hard-concrete $L_0$ gate; \citealp{louizos2018l0}),
initialized open.

\begin{table}[H]
\centering\small
\setlength{\tabcolsep}{3.5pt}
\begin{tabular}{@{}lcccc@{}}
\toprule
 & FourRooms & DMC & Freeway & Synth. \\
Setting & grid & DMC & Atari & synth. \\
\midrule
Observation      & $4{\times}13^2$ & $3{\times}64^2$ & $3{\times}64^2$ & $\mathbb{R}^{64}$ \\
Encoder          & MLP     & CNN     & CNN     & MLP \\
Latent $d$       & $64$    & $256$   & $256$   & $64$ \\
Actions          & 5 disc. & cont.   & 3 disc. & 5 disc. \\
Transitions      & $200$k  & $100$k  & $200$k  & $30$k/$8$k \\
Steps            & $60$k/$100$k$^{*}$ & $50$k   & $40$k  & $5$k \\
Batch            & $256$   & $128$   & $128$   & $256$ \\
Distractor knob  & $n$ cells & $16$ sq. & frameskip & $d_w$ dims \\
\;\;(swept)      & $\{0,10,$ & indep./ & $\{2,4,8\}$ & $\{0,4,$ \\
                 & $30,60\}$ & corr.\  &            & $16,32\}$ \\
\bottomrule
\end{tabular}
\caption{Per-environment settings, for the FourRooms gridworld, DMC
\citep{tassa2018deepmind}, Atari \citep{bellemare2013ale} and the
synthetic testbed. Continuous DMC action dimension is
task-dependent ($1$ cartpole-swingup, $2$ reacher-easy, $6$ cheetah-run);
$n_{\mathrm{cf}}$ counterfactual states $=1000$ (discrete) / $500$
(continuous). Synthetic transitions are $30$k train / $8$k test, self-generated per
cell over $3$ seeds. $^{*}$FourRooms: $60$k steps for the probe-sweep runs,
$100$k for the goal-reaching control runs.}
\label{tab:hparams}
\end{table}

\paragraph{Estimating $\bar\Delta$ for continuous actions.}
$K{=}16$ candidate actions per state: the executed action (index $0$), $7$
actions resampled from the replay buffer, and $8$ Gaussian perturbations of
the executed action with standard deviation $0.3$ of the action half-range,
clipped to the action bounds---the most even split available of the
remaining $K{-}1{=}15$ candidates (the main text's ``split evenly'').
For discrete actions $\bar\Delta$ is the exact
mean over all $|\mathcal{A}|$ offsets.

\paragraph{FourRooms distractors.}
The $13{\times}13$ layout adds a fourth binary observation channel of
$n$ cells that deterministically roll one column per step
(\texttt{np.roll}), independent of the action; we sweep
$n\in\{0,10,30,60\}$. In the action-correlated stress configuration
(\texttt{d30ac}), the same field instead rolls by the agent's own
$(\Delta\mathrm{row},\Delta\mathrm{col})$ displacement each step, so it is no
longer common-mode and centering does not cancel it---the measured boundary.
The five actions are up/down/left/right/no-op.

\paragraph{Synthetic identifiability generator.}
A controllable factor $c\in\mathbb{R}^2$ is moved by fixed per-action
displacements (no-op and $\pm0.3$ along each axis; $5$ actions), clipped to
$[-1,1]$. A distractor $w\in\mathbb{R}^{d_w}$ evolves action-independently as
$w\leftarrow Rw$ with $R$ a fixed block-rotation (angles $\sim
\mathcal{U}(0.2,0.5)$); in the correlated variant $w\leftarrow Rw+V_a$ with
$V_a\sim\mathcal{N}(0,0.15^2)$ per action. The observation is a fixed random
mixing $G$ of $[c;w]$ into $\mathbb{R}^{64}$, either linear (orthonormal
columns) or nonlinear (two \texttt{tanh} layers, width $64$). Episodes have
length $50$; we probe the held-out change feature (the centered offset for
centered/noncentered/gated, the one-step change for standard) against the
true $\Delta c$ ($R^2_c$, recover) and $\Delta w$ ($R^2_w$, reject).

\paragraph{Checkpoint selection.}
The reported ``best'' checkpoint is the lowest validation cosine loss among
checkpoints whose action separation exceeds a regime-aware threshold. An
explicit override always wins; otherwise FourRooms with rolling
action-independent distractors uses threshold $0$---there even healthy runs
sit at AS ${\approx}0.002$ (with offset-probe $R^2\approx0.74$) while dead
ones reach ${\approx}0.008$, so AS carries no collapse signal and the ridge
probe is the health diagnostic---and every other setting uses $0.01$. If no
checkpoint passes, the highest-AS checkpoint is used and the run is flagged
collapsed.

\paragraph{Code and configurations.}
Every model-training run has a named configuration file carrying the
per-experiment overrides above; the synthetic testbeds are self-contained
scripts with their settings inlined. Those configuration files, the run
scripts, and our full training and evaluation code are provided in the
supplementary code archive.

\section{Proof of Proposition~\ref{prop:sample} (Sampled Centering)}
\label{app:proofs}

We restate the proposition, which is proved in full here; it appears as
Proposition~\ref{prop:sample} in the main text.

\setcounter{proposition}{1}
\begin{proposition}[Sampled centering]
Fix $z$ and let
$\bar\Delta_K(z) = \frac{1}{K}\sum_{k=1}^{K}\Delta(z, a_k)$ for candidate
actions $\{a_k\}_{k=1}^{K}$ drawn from a proposal $\nu$, and suppose the
offset head decomposes as $\Delta(z,a) = g(z,a) + h(z)$ for some $g$ and
some action-independent $h$. Then, for every $K$ and every draw:
\begin{enumerate}
  \item[(i)] $\displaystyle \Delta(z,a) - \bar\Delta_K(z)
        = g(z,a) - \frac{1}{K}\sum_{k=1}^{K} g(z,a_k)$, so the common-mode
        component $h$ cancels exactly; and
  \item[(ii)] the sampling error
        $\bar\Delta_K(z) - \mathbb{E}_{a'\sim\nu}[\Delta(z,a')]$ does not
        depend on the query action $a$, is $O(1/\sqrt{K})$ under bounded
        second moments of $g(z,\cdot)$, and cancels in any
        action-differential comparison $u(z,a) - u(z,a')$.
\end{enumerate}
\end{proposition}

\begin{proof}
\emph{Part (i).} Substituting $\Delta(z,a_k) = g(z,a_k) + h(z)$ into the
definition of $\bar\Delta_K$ and using $\frac{1}{K}\sum_{k=1}^{K} 1 = 1$,
\begin{equation*}
\begin{split}
\bar\Delta_K(z)
  &\;=\; \frac{1}{K}\sum_{k=1}^{K}\bigl[g(z,a_k) + h(z)\bigr] \\
  &\;=\; \frac{1}{K}\sum_{k=1}^{K} g(z,a_k) \;+\; h(z),
\end{split}
\end{equation*}
since $h(z)$ is constant across the averaged terms. Subtracting this from
$\Delta(z,a) = g(z,a) + h(z)$ gives
\begin{equation*}
\Delta(z,a) - \bar\Delta_K(z)
  \;=\; g(z,a) - \frac{1}{K}\sum_{k=1}^{K} g(z,a_k),
\end{equation*}
which is claim (i). The cancellation is term by term and therefore holds
for every $K \ge 1$ and every realised candidate set; no independence,
coverage, or unbiasedness property of $\nu$ is used.

\emph{Part (ii).} By the same substitution, and because the $h$ components
coincide in both terms and drop,
\begin{equation*}
\begin{split}
\bar\Delta_K(z) &- \mathbb{E}_{a'\sim\nu}\!\left[\Delta(z,a')\right] \\
  &\;=\; \frac{1}{K}\sum_{k=1}^{K} g(z,a_k)
         - \mathbb{E}_{a'\sim\nu}\!\left[g(z,a')\right].
\end{split}
\end{equation*}
The right-hand side involves only $g$ and is a single quantity that does
not depend on the query action $a$; it is therefore an action-independent
shift of the channel. If the $a_k$ are drawn i.i.d.\ from $\nu$ and
$g(z,\cdot)$ has bounded second moments, it is a sample mean minus its
expectation, so its standard deviation is $O(1/\sqrt{K})$. Finally, since
the shift is common to every action, it cancels in any action-differential
comparison:
\begin{equation*}
\begin{split}
u(z,a) - u(z,a')
  &\;=\; \bigl[\Delta(z,a) - \bar\Delta_K(z)\bigr] \\
  &\qquad - \bigl[\Delta(z,a') - \bar\Delta_K(z)\bigr] \\
  &\;=\; \Delta(z,a) - \Delta(z,a'). \qedhere
\end{split}
\end{equation*}
\end{proof}

Two remarks. \emph{Self-inclusion.} In practice the executed action
occupies one candidate slot ($a_1 = a$); for that query the channel is
attenuated by exactly $(1 - 1/K)$:
$u(z,a) = \frac{K-1}{K}\bigl[g(z,a) -
\frac{1}{K-1}\sum_{k\ge2} g(z,a_k)\bigr]$---an $O(1/K)$ effect on $g$
only, with the $h$-cancellation unaffected. \emph{Proposal bias.} A biased
or narrow $\nu$ (e.g.\ a state-dependent behavior policy) shifts
$\mathbb{E}_{\nu}[g]$ and hence where the channel is centered; by (i) it
cannot re-admit the common mode.
\emph{Empirics.} A $K$-sweep over all $108$ DMC checkpoints
($K\in\{2,4,8,16,32\}$) matches the proposition: channel probe $R^2$ rises
from ${\approx}0$ at $K{=}2$---where the $(1-1/K)$ self-inclusion
attenuation is $\tfrac12$ and the centering point rests on a single fresh
sample---to its plateau by $K\approx16$ (e.g.\ $0.52$ of the $K{=}32$ value
$0.56$ on reacher-independent), while the noncentered channel, which uses
no sampled mean, is exactly flat in $K$---a built-in negative control. The
resampling dispersion of the post-hoc centering point decays monotonically
in $K$ (fitted log-log slope $-0.26\pm0.03$ across noncentered and
standard hosts; shallower than the i.i.d.\ $-\tfrac12$ because the
proposal is structured---the executed action is pinned and the
replay/perturbation composition shifts with $K$). For the \emph{trained-in}
decomposition the candidate-mean prediction is $B(z)$ \emph{exactly}, for
every draw---$\frac1K\sum_k [\Delta(z,a_k)-\bar\Delta_K(z)] = 0$---so its
output's centering point is draw-invariant by construction (measured
dispersion at float precision, ${\sim}10^{-7}$): a small but provable
advantage of embedding the readout in the parameterization.

\section{The Out-of-Distribution Goal Pathology}
\label{app:goalfix}
Episodes that terminate on goal contact leave the on-goal state absent from
the training marginal: in 200k FourRooms transitions it never appears as a
start state. The converged encoder then decodes the on-goal state
$10.5$~cells off (all other cells: ${\sim}0.005$), so no imagined plan can
ever terminate ``at the goal'' and MPC success is $0.00$ despite perfect
in-distribution prediction, rollout, and position-decoding metrics ---
metrics that are all blind to the missing state. Allowing the collection
policy (only) to walk through the goal restores the state to the training
distribution and MPC success to $1.00$ across both variants and all seeds,
with in-distribution metrics unchanged. We report this as a cautionary
protocol note: rollout fidelity does not imply planning success when the
plan's target is itself out of distribution.

\section{Distracting DMC: Pre-Registered Gate --- Criteria and Outcomes}
\label{app:e3gate}
The DMC decision criteria were fixed before the matrix ran: (i) a
centered-minus-noncentered probe gap $\ge 0.2$ under independent
distraction with 3/3 seed sign-consistency per task; (ii) an absolute bar
of $R^2 \ge 0.5$ under distraction; (iii) a correlated-regime gap within
$\pm 0.05$. Outcomes, verbatim: (i) \textbf{pass} on 3/3 tasks, 9/9 seeds
(gaps $+0.212$/$+0.244$/$+0.493$ on cheetah/cartpole/reacher); (ii)
\textbf{fail}, diagnosed as ceiling mis-calibration---the distraction-free
probe ceilings are $0.26$ (cartpole) and ${\approx}0.32$--$0.40$ (cheetah),
so a fixed absolute bar cannot be met even without distractors on two of
three tasks (only reacher's ceiling exceeds $0.5$); (iii) within tolerance
on 2/3 tasks ($+0.033$ cheetah, $-0.040$ cartpole), with the reacher
inversion ($-0.153$) analyzed as contamination-boost: the noncentered
offset's correlated-regime probe ($0.535$) exceeds its own
distraction-free level ($0.169$), so the surplus is distractor signal
masquerading as action effect. Per-seed values for the four
variants run at DMC scale---standard (which exposes no isolated channel),
noncentered, centered, and gated (whose gate is inert)---are given in
Table~\ref{tab:e3perseed}; the \emph{residual} rung of the main text's
ladder was not part of the DMC matrix.

\section{Plug-in: Per-Cell Numbers}
\label{app:plugin}
Table~\ref{tab:plugin} gives the exact per-cell probe values behind
Fig.~\ref{fig:pluginbar}. Readout variance over five probe seeds (same
checkpoints, resampled evaluation data and probe splits): the centered
column varies by s.d.\ $0.009$--$0.030$ per cell (RePo
$0.260{\pm}.024$/$0.300{\pm}.016$/$0.454{\pm}.014$, TIA
$0.184{\pm}.030$/$0.336{\pm}.009$/$0.504{\pm}.011$ for
cartpole/cheetah/reacher), with raw $\Delta$ at $\le 0.061$ throughout;
the one-training-run caveat is a host property, not readout noise.

\begin{table}[H]
\centering\small
\setlength{\tabcolsep}{4pt}
\begin{tabular}{@{}llcccc@{}}
\toprule
host & task & full latent & raw $\Delta$ & centered & trained-in \\
\midrule
RePo & cartpole & $0.559$ & $-0.002$ & $\mathbf{0.264}$ & $0.268$ \\
     & cheetah  & $0.644$ & $0.026$  & $\mathbf{0.323}$ & $0.349$ \\
     & reacher  & $0.624$ & $-0.002$ & $\mathbf{0.457}$ & $0.577$ \\
\midrule
TIA  & cartpole & $0.027$ & $-0.002$ & $\mathbf{0.190}$ & $0.198$ \\
     & cheetah  & $0.789$ & $0.052$  & $\mathbf{0.345}$ & $0.168$ \\
     & reacher  & $0.737$ & $0.006$  & $\mathbf{0.519}$ & $0.506$ \\
\bottomrule
\end{tabular}
\caption{\textbf{Centering as a plug-in} (probe $R^2$, higher is better
except raw $\Delta$ which diagnoses the entangled readout).
RePo \citep{zhu2023repo} and TIA \citep{fu2021tia}
trained 500k steps under DAVIS video backgrounds
\citep{ponttuset2017davis}, one training seed per
cell; probes on frozen final checkpoints over $5{,}000$ held-out
random-policy transitions, target = change in physics state. \emph{Raw
$\Delta$} vs.\ \emph{centered}: identical model, identical data---the only
change is subtracting the $K{=}16$ action-mean at readout.
\emph{Trained-in}: centering inside the RSSM prior
\citep{hafner2019planet} for the full 500k run
(both hosts). Budgets and protocols differ from ours; no cross-method
return comparison is implied.}
\label{tab:plugin}
\end{table}

\section{The Gate Ablation Is Inert}
\label{app:gateablation}

The gated variant was designed as a learned per-dimension controllability
mask: a sigmoid gate $C(z)\in(0,1)^d$ under an L1 sparsity penalty,
intended to select which latent dimensions the action channel may write to.
It does not learn this. Across all testbeds the gate converges to
near-uniform down-scaling rather than sparse selection---an inert
reparameterization, since a uniform gate can be absorbed into the scale of
$\Delta$, and L1-on-sigmoid admits exactly this shrink-everything solution.
Replacing L1 with a hard-concrete $L_0$ relaxation
\citep{louizos2018l0} does not rescue it: the $L_0$ gate collapses at
every sparsity coefficient we tested. The gate does not improve channel
purity over plain centering and adds a genuine failure mode at the strongest
distractor setting (Sec.~\ref{sec:e6}). We therefore report it as a negative
result: AD-JEPA sets $C\equiv 1$, centering is the sole mechanism, and the
gated rung survives only as an ablation.

\section{Steering Along the Channel Under Distraction}
\label{app:steer}

We test the action channel by \emph{perturbation} rather than by
correlation. For
a frozen DMC checkpoint we push the latent along the model's own channel
$u(z,a)$ (centered offset for centered/gated, raw offset for noncentered,
$\hat z' - z$ for standard), project back to the encoder's layer-norm
manifold, and decode the induced state displacement with a nonlinear MLP
trained only as a measurement instrument. The steering direction comes
from the channel alone---no privileged signal enters, unlike the
compass-scored planner of Appendix~\ref{app:discussion}. We report the cosine
between the decoded displacement and the action's counterfactual state
effect $\Delta s(a) - \frac{1}{K}\sum_{k}\Delta s(a_k)$ (cf-cos) and, as
the load-bearing control, the same quantity for an \emph{equal-norm random
push} through the identical decoder---absolute decoder quality cancels, so
the comparison is instrument-relative. $3$ tasks $\times$ $3$ distractor
modes $\times$ $4$ variants $\times$ $3$ seeds; push scales
$\gamma\in\{0.5,1,2\}$; we quote $\gamma{=}1$ throughout (cf-cos is
positive and monotone in $\gamma$ wherever it is nonzero, so no push scale
is selected post hoc). Pushes are first-order and the predictor is not
rolled forward from the pushed latent: the measurement is
directional---channel alignment, not closed-loop control.

\paragraph{Result (cheetah-run).} Without distractors every channel
steers (cf-cos $0.42/0.42/0.29/0.43$ for centered/gated/noncentered/
standard). Under \emph{independent} distraction only the centered variants
survive ($0.27/0.26$ vs.\ $0.00/0.02$; random control ${\approx}0.01$
throughout; $3/3$ seeds): centering retains ${\sim}65\%$ of its
no-distractor steering where the raw-offset and standard channels retain
essentially none. Two properties sharpen this. \emph{(i) Norm does not buy
efficacy}: the channel that steers has $\lVert u\rVert\approx0.13$; the
standard channel that does not has $\lVert u\rVert\approx16$. \emph{(ii)
Selectivity}: a centered push moves the decoded \emph{distractor} state
about half as much as an equal-norm random push, and about $5\times$ less
than the noncentered channel's push. Under \emph{action-correlated}
distraction centered steering collapses to ${\approx}0.07$ even at the
most favorable push scale---the pre-registered boundary.

\paragraph{Scope and instrument.} The instrument is a nonlinear MLP;
because the matched random push absorbs decoder quality, the claim is
directional and instrument-relative---what a low absolute decode $R^2$
under distraction permits. The contrast is not an instrument artifact:
across five decoder configurations (widths $128$--$512$, depths $1$--$3$,
with and without input noise; test $R^2$ spanning $-1.8$ to $-3.6$), the
centered channel's alignment is flat ($0.25$--$0.31$) while the
noncentered and standard channels stay at ${\approx}0$ in every
configuration. reacher-easy is uninformative: no channel
steers even without distractors, consistent with its documented
undertraining at this budget (Appendix~\ref{app:e3gate}) rather than a task
property---fully trained baselines reach state $R^2$ $0.80$--$0.92$ on the
same task, and post-hoc centering on those frozen models yields a
$0.46$--$0.58$ channel (Sec.~\ref{sec:plugin}). cartpole-swingup is weakly
positive for the centered variant only ($0.06$ vs.\ random $0.02$). We
therefore report cheetah-run as the clean case and the other two as gated
out by a no-distractor model-quality check, per our report-either-way
pre-commitment.

\subsection{Frozen Third-Party Hosts}
\label{app:hoststeer}

The same intervention ports to the frozen RePo/TIA hosts of
Sec.~\ref{sec:plugin}---RSSM architectures and objectives we did not
design, trained with the authors' released code and frozen before the
readout is applied. These hosts run under DAVIS video (action-independent
distraction only, no distractor ground truth), so the test measures
\emph{action-specificity} of the channel direction, not common-mode
rejection. We push the host's stochastic state along the post-hoc channel
$u(z,a) = m(z,a) - \frac{1}{16}\sum_{k} m(z,a_k)$ of the RSSM prior mean
$m$, decode with an MLP fitted per host (shuffled-split calibration,
matching the reference ridge protocol), and score the decoded displacement
against the action's counterfactual-centered physics effect at fixed
$\gamma{=}1$; three evaluation seeds per host, one training run per cell.

\begin{table}[H]
\centering\small
\setlength{\tabcolsep}{3.2pt}
\begin{tabular}{@{}llcccccc@{}}
\toprule
host & task & centered & glob.\ & raw & rand & perm & $R^2$/$R^2_{s}$ \\
\midrule
RePo & cartpole & $\mathbf{+.41{\pm}.08}$ & $+.02$ & $+.02$ & $+.02$ & $+.01$ & $.85$/$.32$ \\
RePo & cheetah  & $\mathbf{+.09{\pm}.02}$ & $-.00$ & $+.00$ & $+.00$ & $-.00$ & $.76$/$.11$ \\
RePo & reacher  & $-.02{\pm}.05$ & $-.01$ & $-.01$ & $+.01$ & $-.02$ & $.88$/$.04$ \\
TIA  & cheetah  & $\mathbf{+.14{\pm}.01}$ & $+.05$ & $+.04$ & $-.01$ & $+.01$ & $.87$/$.20$ \\
TIA  & reacher  & $\mathbf{+.18{\pm}.05}$ & $+.06$ & $+.07$ & $-.00$ & $-.01$ & $.97$/$.39$ \\
\bottomrule
\end{tabular}
\caption{\textbf{Steering frozen third-party hosts} (cosine to the
counterfactual-centered physics effect, $\gamma{=}1$, mean$\pm$s.d.\ over
3 evaluation seeds). Controls: \emph{glob.}\ subtracts the global mean
offset instead of the state-conditional marginal; \emph{rand} is an
equal-norm random push through the same decoder; \emph{perm} scores
against permuted (deranged) targets. $R^2$/$R^2_s$: decoder fit from the
full latent / from the pushed state block alone. TIA-cartpole is excluded:
its instrument fails ($R^2{=}-0.07$) on a host whose full latent is
independently known to be collapsed ($0.03$, Table~\ref{tab:plugin}).}
\label{tab:hoststeer}
\end{table}

Four observations. \emph{(i)} The three controls isolate the channel: the
equal-norm random push and the permutation null are ${\approx}0$
everywhere (the alignment carries per-sample action information), and the
global-mean control tracks raw rather than centered---only the
state-conditional marginal recovers the channel, the observation behind
the design note in Sec.~\ref{sec:method}. \emph{(ii)} repo-reacher is an
honest null on a validated instrument ($R^2{=}0.88$): that host keeps
physics almost entirely in its deterministic belief ($R^2_{\text{belief}}
= 0.89$ vs.\ $R^2_s = 0.04$), so the pushed block has nothing to steer;
across the five validated cells the alignment tracks $R^2_s$ (Spearman
$\rho = 0.90$, $n{=}5$---suggestive, not confirmatory). \emph{(iii)}
Stability tracks instrument validity: across three independent evaluation
runs, validated-instrument cells replicate closely (repo-cartpole moves by
$0.003$) while broken-instrument numbers moved arbitrarily in both
directions---we therefore report only validated cells. \emph{(iv)} The
result is directional (first-order pushes; the host is never rolled
forward) and says nothing about distractor invariance, which DAVIS hosts
cannot measure.

\section{Extended Discussion}
\label{app:discussion}

This appendix gives the full-length analysis of the control null
summarized in Sec.~\ref{sec:limitations}.

\paragraph{The one-step predictor does not use the channel unaided.}
Action separation (AS; the mean pairwise distance between per-action
predictions) collapses from ${\sim}1.28$ to ${\sim}0.002$ for \emph{every}
variant---including centered---once FourRooms distraction reaches
$n \ge 30$ cells, even though the centered offset probe still reads the
agent factor at $R^2=0.73$ from the same checkpoints
(Sec.~\ref{sec:e1}). The channel is preserved but unused: when distractor
variance dominates the prediction target, the one-step objective is nearly
indifferent to the action, and the predictor shrinks its action sensitivity
toward zero. A consumer with a longer horizon can restore it at small
scale: in the FourRooms grid, the gridworld planner, trained with a multi-step
rollout loss, reaches $0.92\pm0.02$ goal-reaching success at $n{=}30$ where
the standard predictor manages $0.57\pm0.03$ (Sec.~\ref{sec:e2}); this
conversion does not replicate at DMC scale (below). The practical
corollary is the AS-gated checkpoint-selection rule of
Sec.~\ref{sec:protocol}; we suspect the same selection pathology affects
other latent world models trained under heavy distraction.

\paragraph{The longer-horizon consumer does not restore control at DMC
scale.}
The multi-step planner converts the preserved channel into behavior in
FourRooms, but the conversion does not replicate at DMC scale within our
compute budget. We fixed a pre-registered gate---beat a random-action
reference at goal tolerances $0.05/0.1$---before running, ported the gridworld
recipe to reacher-easy (three seeds per variant trained to 200k steps with
the same multi-step rollout loss, planned with CEM MPC
\citep{deboer2005cem} at 30 episodes per
run, five of the six runs evaluated, and scored by the
same position-decoding compass head the gridworld planner uses---privileged, and
used only to score candidate plans), and missed the gate across the full
effort ladder. The outcome is not merely at-chance: at the tolerances
where a random policy has traction, the centered planner is
\emph{sub-random} (goal-reaching success $0.03/0.10$ vs.\ $0.23/0.57$ at
tolerance $0.2/0.3$, two seeds)---planning through the model is worse
there than not planning---while at the pre-registered tight tolerances
both planner and random are at zero, and the aggregate final distance
matches random ($2.35$ vs.\ $2.44$). The standard ablation is strictly
worse: zero success at \emph{every} tolerance across all three of its
seeds ($0/90$ episodes) at mean final distance $4.64$. The null holds
across $4\times$ training ($50\mathrm{k}\!\to\!200\mathrm{k}$ steps),
$32\times$ planning budget (shipped$\to$huge CEM), the multi-step rollout
loss, and oracle scoring; on the earlier 50k checkpoints performance
additionally \emph{degraded} as the planning horizon grew---the signature
of search through a drifting model. Because the compass scorer is
privileged---an upper bound on the scoring information any reward-free
readout could supply---the null is generous-case: the planner lost even
with the best available plan scorer, which forecloses a ``needs a better
reward'' rescue.

\paragraph{The null is consistent with a model-accuracy bottleneck, not a
channel defect.}
Within an episode the centered planner transiently steers the right
way---best in-episode distance falls to ${\sim}0.15$---before drifting
back out to final distances of ${\sim}1$--$5$: the expected signature of
CEM planning through an inaccurate model, not of a channel that
misdirects. Consistent with a representation that is not itself the
bottleneck, the centered readout is the more accurate open-loop model:
five-step decoded-state rollout error (standardized) is $0.62$ for
centered vs.\ $0.74$ for standard, non-overlapping across three seeds; but
both sit far above the sub-$0.4$ target we pre-set when porting the
recipe, and both drift upward late in training. We therefore read the null
as a \emph{candidate} model-accuracy limit rather than a proven one. A
direct intervention on the same frozen checkpoints corroborates this
localization: pushing the latent along the model's own centered channel
reproduces the executed action's counterfactual state effect on
cheetah-run under independent distraction (cf-cos ${\approx}0.27$ at
$\gamma{=}1$ vs.\ ${\approx}0.01$ for an equal-norm random push through
the same decoder), where the noncentered and standard channels---which
steer comparably \emph{without} distractors---collapse to ${\approx}0$;
the effect is monotone in the push scale and respects the
action-correlated boundary (Appendix~\ref{app:steer}). The failure is also
distinct from the out-of-distribution-goal artifact of
Appendix~\ref{app:goalfix}: there the goal state was absent from the training
marginal; here the goal is in-distribution and the rollout itself drifts.

\section{Scope of the Claim}
\label{app:scope}

\paragraph{What the identity does---and does not---guarantee.}
Proposition~\ref{prop:cancel} removes whatever the \emph{learned offsets}
represent identically across actions; it does not by itself guarantee that
all action-independent distractors are removed. If the offset head learns
a zero-mean action interaction, $\Delta(z,a) = h(z) +
s_{\mathrm{dist}}(z)\,\phi(a)$ with $\mathbb{E}_{a'\sim\nu}\,\phi(a')=0$,
centering leaves $s_{\mathrm{dist}}(z)\,\phi(a)$ intact even though the
distractor's environmental dynamics never depend on the action. Invariance
to a distractor is therefore a joint property of the identity and of what
the trained offsets happen to represent---which is precisely what the
channel-purity probes measure rather than assume. Empirically, such
interactions do not arise where the nuisance is action-independent:
centered-channel leak is indistinguishable from zero (Secs.~\ref{sec:e4},
\ref{sec:e6}).

\paragraph{Sufficiency, not superiority.}
The representation-to-control conversion is demonstrated in the
$13\times13$ FourRooms gridworld (centered $0.92$ vs.\ standard $0.57$ at
$n{=}30$; Sec.~\ref{sec:e2}) and does not replicate at DMC scale at this
compute budget; we treat it as a bounded existence proof that the
preserved channel \emph{can} convert to control where the world model is
accurate enough to plan through---which our gridworld models are and our
continuous-control models, at this budget, are not---and never as a
control-superiority claim at scale.

\paragraph{Single-agent scope.}
Centering marginalizes over the ego agent's action alone. Entities driven
by their own policies cancel only while their transitions are common-mode
with respect to that action; the moment they react to the
agent---pursuit, evasion, coordination---they become action-correlated and
inherit the reafferent boundary. A multi-agent extension would center over
joint actions or per-agent marginals, which presupposes observing or
inferring the other agents' actions---exactly the supervision our
action-labeled setting assumes only for the ego agent.

\section{Supplementary Figures and Tables}
\label{app:floats}

This section collects the floats relocated from the main
text for the page limit, in ascending float number. Each is
reproduced exactly as generated; the surrounding text says only what the float
contains and how to read it.

We begin with the two natural-distraction testbeds. Table~\ref{tab:e6_freeway}
gives the Freeway channel-purity numbers behind the discussion in the main
text: the agent's own displacement read out of the effective offset, scored
against the one-hot(action) ceiling, alongside the traffic leak measured
against both the traffic change and the traffic state.

\begin{table}[H]
\centering
\setlength{\tabcolsep}{2pt}\scriptsize
\begin{tabular}{@{}lccccc@{}}
\toprule
 & \multicolumn{3}{c}{agent own-$\Delta$ $R^2\uparrow$} & traffic $R^2\downarrow$ & separation$\uparrow$ \\
\cmidrule(lr){2-4}
features & fs\,2 & fs\,4 & fs\,8 & ($\Delta$ \& state) & (mean$\pm$s.d.) \\
\midrule
one-hot$(a)$ ceiling & 0.810 & 0.757 & 0.683 & --- & --- \\
\midrule
\textbf{centered} (ours) & \textbf{0.877} & \textbf{0.838} & \textbf{0.795} & $[-0.001,0.000]$ & $\mathbf{+0.837{\pm}0.036}$ \\
\; excess$^{\dagger}$ & \multicolumn{3}{c}{$+0.19$--$0.23$ (9/9)} & & \\
gated & 0.825 & 0.851 & 0.788$^{\ddagger}$ & $[-0.001,0.000]$ & $+0.736\pm0.272$ \\
noncentered & \multicolumn{3}{c}{$0.149$--$0.485$} & $0.545$--$0.612$ & $-0.256\pm0.302$ \\
\midrule
full latent $z$ (any var.) & \multicolumn{3}{c}{---} & $\approx 0.87$ & --- \\
\bottomrule
\end{tabular}
\caption{\textbf{Freeway channel purity} (end-of-training checkpoints,
3 seeds per frameskip; the 3/36 collapsed runs are kept in pooled
separation statistics, excluded from per-frameskip aggregates). Ridge
$R^2$ from the effective offset for the agent's displacement (read
against the one-hot ceiling) and the traffic factors; per the predictability control the leak
column covers traffic \emph{state} as well as $\Delta$. Separation $=$
agent $R^2$ $-$ traffic-state $R^2$ per run.
$^{\dagger}$Excess $=$ agent $R^2$ after regressing out one-hot$(a)$.
$^{\ddagger}$Gated fs\,8: mean of the two healthy seeds.
Bottom row: traffic state decoded from the \emph{full latent} $z$ rather
than the offset---rejection is routing, not blindness.}
\label{tab:e6_freeway}
\end{table}
Table~\ref{tab:e3dmc} is the continuous-control counterpart,
covering three DMC tasks under the three distractor regimes. The independent
column is the one the claim rests on; the correlated column is the stated
boundary, and reacher-easy's noncentered cell there is the
contamination-as-signal case discussed above.

\begin{table}[H]
\centering\small
\setlength{\tabcolsep}{4.5pt}
\begin{tabular}{@{}llccc@{}}
\toprule
 & & \multicolumn{3}{c}{probe $R^2$ (effective offset)} \\
\cmidrule(lr){3-5}
task & method & none & indep. & corr. \\
\midrule
reacher-easy & noncentered & $0.169$ & $0.039$ & $0.535$ \\
 & centered (ours) & $0.162$ & $\mathbf{0.533}$ & $0.382$ \\
\midrule
cheetah-run & noncentered & $0.398$ & $0.128$ & $0.223$ \\
 & centered (ours) & $0.321$ & $\mathbf{0.340}$ & $0.256$ \\
\midrule
cartpole-swingup & noncentered & $0.116$ & $0.003$ & $0.279$ \\
 & centered (ours) & $0.261$ & $\mathbf{0.248}$ & $0.239$ \\
\bottomrule
\end{tabular}
\caption{\textbf{Distracting DMC} (3 seeds; per-seed values and seed
s.d.\ in Table~\ref{tab:e3perseed}---s.d.\ $\le 0.014$ outside
reacher-easy, whose correlated cells reach $0.094$). Probe $R^2$ of the
controllable state change from the effective offset. Under independent
distraction (bold) the centered channel survives on all tasks (9/9
seeds); the correlated column is the stated boundary, and reacher's
noncentered correlated cell is contamination-as-signal.}
\label{tab:e3dmc}
\end{table}

\begin{figure}[H]
\centering
\includegraphics[width=\columnwidth]{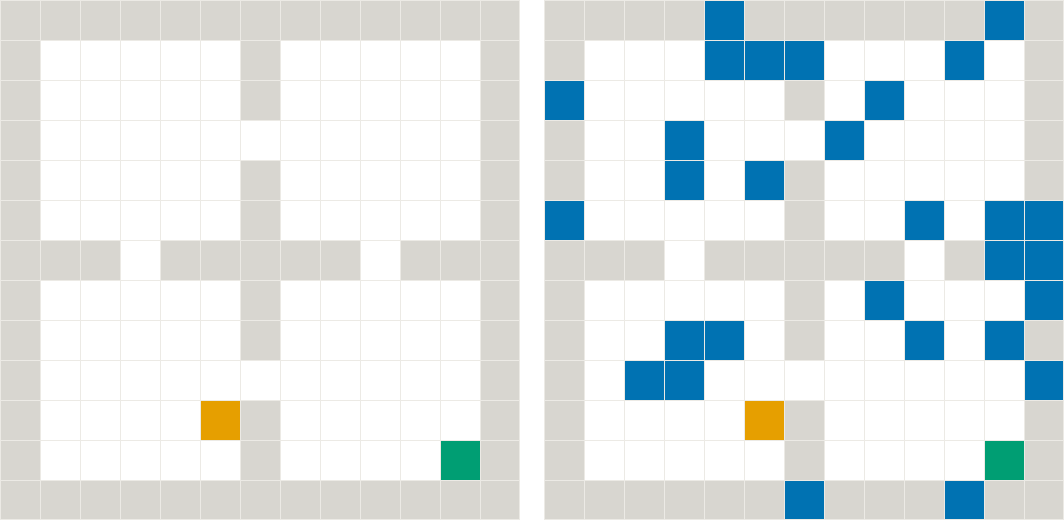}
\caption{\textbf{What ``distraction'' is, concretely} (FourRooms
observations, rendered from the environment). \emph{Left:} $n{=}0$.
\emph{Right:} $n{=}30$ rolling distractor cells (blue). The room, the agent
(orange) and the goal (green) are identical; only the distractor field
differs, and it rolls one column per step \emph{regardless of the action}.
That per-transition invariance across actions is what the action mean
absorbs. In the action-correlated configuration the same field instead rolls
by the agent's own displacement, which is the boundary of the guarantee.}
\label{fig:frames}
\end{figure}

\begin{figure}[H]
\centering
\includegraphics[width=\columnwidth]{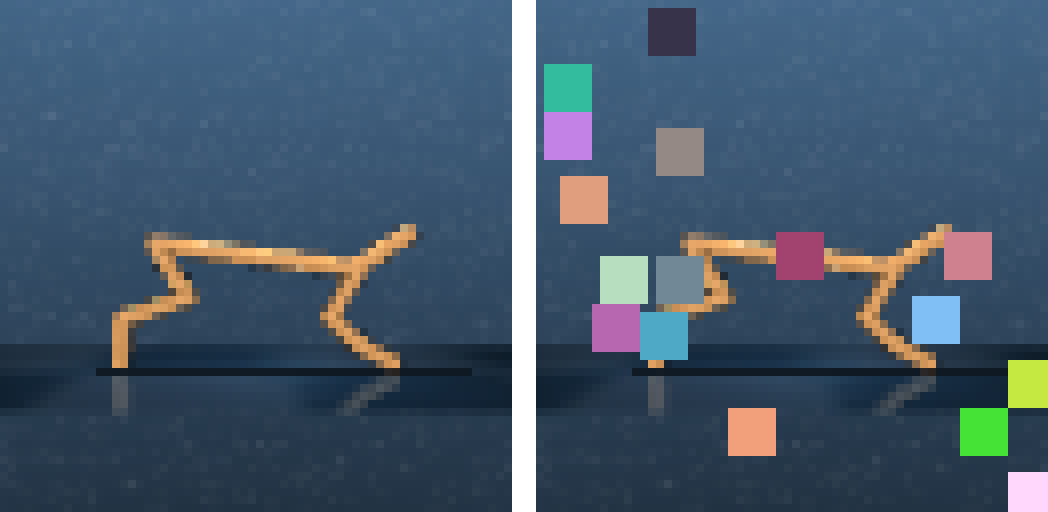}
\caption{\textbf{The same, in continuous control} (cheetah-run
observations, rendered from the environment at the settings every DMC run
used: $64{\times}64$, $16$ occluders). \emph{Left:} clean.
\emph{Right:} the same transition under distraction. Both panels are
stepped with the same action sequence and the occluders are composited
onto the pixels only, so the physics---and with it the probe target---is
bit-identical between them while $13.6\%$ of pixels change. In the
independent configuration the occluders translate a fixed $2$\,px per
frame whatever the action; in the action-correlated configuration they
translate by the action instead. A single frame cannot tell the two apart:
they differ only in the motion rule, which is precisely the property the
action mean is sensitive to.}
\label{fig:dmcframes}
\end{figure}

Figure~\ref{fig:dmcframes} shows the same contrast for
continuous control, at the exact settings every DMC run used.

The next group covers the synthetic generator, where the
controllable and nuisance factors are known by construction and both halves of
the claim can be scored directly. Table~\ref{tab:e4} summarises recovery and
rejection across the full grid.

\begin{table}[H]
\centering
\small
\begin{tabular}{lcc}
\toprule
Variant & $R^2_{\mathrm{c}}$ (recover $\uparrow$) & $R^2_{\mathrm{w}}$ (reject $\downarrow$) \\
\midrule
Standard        & 0.02--0.04$^{\dagger}$ & ---           \\
Noncentered     & 0.75--0.89             & 0.74--0.97    \\
Centered (ours)  & \textbf{0.87--0.93}    & $\mathbf{-0.00}$ \\
\bottomrule
\end{tabular}
\caption{Synthetic identifiability: ranges of held-out probe $R^2$ across the
120-cell grid (3 seeds per cell). $^{\dagger}$At nuisance dimension
$d_{\mathrm{w}}\ge 16$; recovery fails outright, so rejection is moot
(cell not reported). $-0.00$ denotes a value that is negative and rounds to
zero; held-out $R^2$ can be slightly negative.}
\label{tab:e4}
\end{table}

Table~\ref{tab:e1} gives the gridworld sweep the main text quotes,
including the action-correlated configuration at which both variants fail.

\begin{table}[H]
\centering
\small
\begin{tabular}{lccccc}
\toprule
& \multicolumn{4}{c}{rolling distractor cells $n$} & \\
\cmidrule(lr){2-5}
Offset probe $R^2$ & 0 & 10 & 30 & 60 & d30ac \\
\midrule
Centered (ours) & 0.31 & 0.55 & \textbf{0.73} & \textbf{0.69} & $-0.00$ \\
Noncentered    & 0.18 & 0.21 & 0.00 & $-0.00$ & $-0.00$ \\
\bottomrule
\end{tabular}
\caption{\textbf{Gridworld distractor sweep} (held-out probe $R^2$, agent factor from the
effective offset; 3 seeds). Both variants die under action-correlated
d30ac---the stated boundary.}
\label{tab:e1}
\end{table}

\begin{figure}[H]
\centering
\includegraphics[width=\columnwidth]{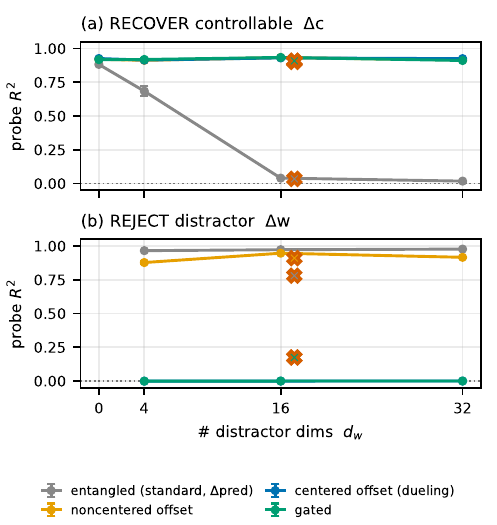}
\caption{Synthetic identifiability on the \emph{linear} generator:
recover ($R^2_{\mathrm{c}}$, top) versus reject ($R^2_{\mathrm{w}}$, bottom)
against the nuisance dimension $d_w$. The $\times$ markers are the
action-correlated boundary at $d_w{=}16$. Error bars: s.e.m.\ over 3
seeds. Values are discussed in the text;
the nonlinear generator and the full $120$-cell grid are in
Table~\ref{tab:e4}.}
\label{fig:e4}
\end{figure}

Figure~\ref{fig:e4} plots the same synthetic study against the
nuisance dimension, so the two requirements can be read against each other as
the distractor grows. The curves sweep $d_w$ on the \emph{linear} generator;
the $\times$ markers, offset slightly to the right of $d_w{=}16$, carry the
action-correlated configuration at that same dimension, so the boundary can be
compared against the action-independent case without a second panel. Reading
the two panels together separates the two ways a channel can fail. Panel~(a)
is the recovery requirement: both centered variants hold at $0.91$--$0.93$
across the sweep, and the noncentered ablation recovers just as well, so
recovery alone does not discriminate between them. Panel~(b) is the rejection
requirement, and there the two separate completely---the centered variants sit
at $-0.00$ at every dimension while the noncentered offset carries the
distractor at $0.88$--$0.95$. The standard predictor fails the first
requirement outright rather than the second: its recovery decays from $0.88$
to $0.02$ as $d_w$ grows, which is why its rejection column is not
informative and is left unreported in Table~\ref{tab:e4}. At the boundary
markers, rejection breaks for the centered variants as well
($-0.00 \to 0.17$), which is the same failure the gridworld sweep shows under
its action-correlated configuration.

Two controls follow. Table~\ref{tab:e4n} varies how predictable
the nuisance dynamics are, separating a leak of the nuisance \emph{change} from
a leak of the nuisance \emph{state}---the distinction that makes the leak
metric informative when the nuisance is stochastic.

\begin{table}[H]
\centering
\small
\begin{tabular}{llccc}
\toprule
 & & \multicolumn{3}{c}{nuisance predictability} \\
\cmidrule(lr){3-5}
Variant & Leak metric & $1.00$ & $0.76$ & $0.05$ \\
\midrule
Noncentered & $\Delta_{\mathrm{w}}$-leak & 0.947 & 0.578 & 0.022 \\
Noncentered & state-leak                 & 0.947 & 0.797 & 0.804 \\
Standard    & state-leak                 & 0.97  & 0.89  & 0.97  \\
Centered (ours) & state-leak & \multicolumn{3}{c}{$-0.01$ to $-0.00$ (all levels)} \\
\bottomrule
\end{tabular}
\caption{Predictability control: leak of the action channel under increasingly stochastic
nuisance dynamics. $\Delta$-leak vanishes for the noncentered ablation as
the nuisance becomes unpredictable, but its state-leak persists: the channel
still encodes where the nuisance \emph{is}. The centered variant is clean at
every predictability level. Entries are means over 3 seeds.}
\label{tab:e4n}
\end{table}

Table~\ref{tab:stage0} repeats the recover-and-reject test in
the continuous-action regime, where the action mean is a Monte-Carlo estimate
rather than an exact enumeration.

\begin{table}[H]
\centering
\small
\begin{tabular}{lcc}
\toprule
Variant & $R^2_{\mathrm{c}} \uparrow$ & $R^2_{\mathrm{w}} \downarrow$ \\
\midrule
Standard       & 0.04 & 0.97 \\
Noncentered    & 0.81 & 0.96 \\
Centered (ours) & \textbf{0.87} & $\mathbf{-0.00}$ \\
\midrule
Centered, action-\emph{correlated} nuisance & --- & 0.19 \\
\bottomrule
\end{tabular}
\caption{Stage 0 (continuous actions): sampled centering
($K{=}16$; entries are means over 3 seeds). The discrete-case pattern survives Monte-Carlo
estimation of $\bar\Delta$; an action-correlated nuisance breaks rejection,
marking the common-mode boundary rather than an estimator failure.}
\label{tab:stage0}
\end{table}

The remaining floats concern behaviour rather than
representation. Table~\ref{tab:e2} reports goal-reaching success under latent
MPC across the gridworld sweep.

\begin{table}[H]
\centering
\small
\setlength{\tabcolsep}{4pt}
\begin{tabular}{lcccc}
\toprule
& \multicolumn{4}{c}{rolling distractor cells $n$} \\
\cmidrule(lr){2-5}
Success rate & 0 & 10 & 30 & 60 \\
\midrule
Standard        & 0.98 & 0.72$\pm$0.07 & 0.57$\pm$0.03 & 0.71$\pm$0.10 \\
Centered (ours)  & 1.00 & 0.93$\pm$0.01 & 0.92$\pm$0.02 & 0.93$\pm$0.01 \\
Gated           & 1.00 & 0.99$\pm$0.01 & 0.99$\pm$0.01 & 0.95$\pm$0.02 \\
\midrule
Random policy   & \multicolumn{4}{c}{0.18--0.28} \\
\bottomrule
\end{tabular}
\caption{\textbf{Gridworld control: the planner converts the preserved channel into
control.} Goal-reaching success under latent MPC (FourRooms; variants
run here: standard, centered, gated; 3 seeds, mean$\pm$s.e.m.). Both
centered variants stay near ceiling across the sweep while the standard
predictor degrades. Under the action-correlated d30ac configuration, all
three variants fall to $0.55$--$0.57$ --- the same boundary as the probe
sweep.}
\label{tab:e2}
\end{table}

\begin{figure}[H]
\centering
\includegraphics[width=\columnwidth]{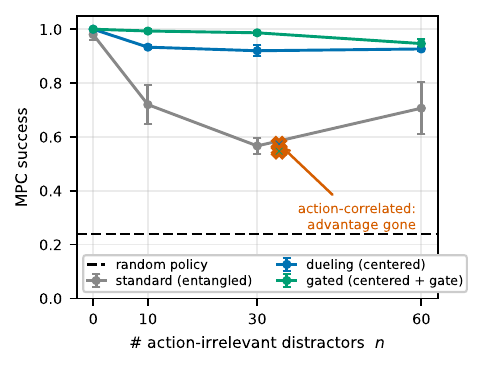}
\caption{\textbf{Gridworld control across the distractor sweep.} Decomposed variants hold
near ceiling where the standard predictor degrades; representation and
control break at the same action-correlated point. Variants shown:
standard, centered, gated; error bars are s.e.m.\ over 3 seeds. Legends
use \emph{dueling} for centered.}
\label{fig:twin}

\end{figure}

Figure~\ref{fig:twin} places the probe and control curves side
by side, so the representation boundary and the control boundary can be
compared directly.

\begin{figure}[H]
\centering
\includegraphics[width=\columnwidth]{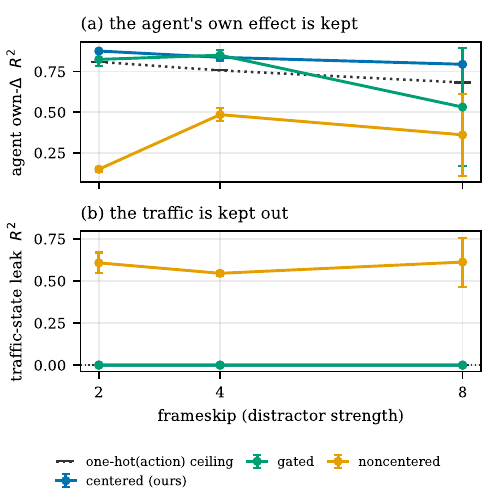}
\caption{\textbf{Freeway.} Agent vs.\ traffic probe $R^2$ from the
effective offset across frameskips (dotted $=$ one-hot ceiling): centered
decodes the agent above the ceiling with traffic leak pinned at zero;
noncentered inverts the pattern. Error bars: s.d.\ over 3 seeds.}
\label{fig:e6_atari}
\end{figure}

Figure~\ref{fig:e6_atari} shows the Freeway readout across
frameskips in the same format.

Finally, Table~\ref{tab:e3perseed} lists every DMC run
individually, so the per-seed spread behind the aggregated cells above can be
inspected directly.

\begin{table}[H]
\centering
\setlength{\tabcolsep}{3pt}\scriptsize
\begin{tabular}{@{}llccccc@{}}
\toprule
variant & regime & $s_0$ & $s_1$ & $s_2$ & mean & s.d. \\
\midrule
\multicolumn{7}{@{}l}{\emph{reacher-easy}}\\[1pt]
noncentered & none & $0.183$ & $0.158$ & $0.167$ & $0.169$ & $0.013$ \\
 & indep. & $0.046$ & $0.040$ & $0.032$ & $0.039$ & $0.007$ \\
 & corr. & $0.576$ & $0.428$ & $0.603$ & $0.535$ & $0.094$ \\
centered (ours) & none & $0.147$ & $0.142$ & $0.197$ & $0.162$ & $0.030$ \\
 & indep. & $0.530$ & $0.536$ & $0.532$ & $0.533$ & $0.003$ \\
 & corr. & $0.459$ & $0.336$ & $0.351$ & $0.382$ & $0.067$ \\
gated & none & $0.113$ & $0.024$ & $0.115$ & $0.084$ & $0.052$ \\
 & indep. & $0.528$ & $0.534$ & $0.526$ & $0.529$ & $0.004$ \\
 & corr. & $0.287$ & $0.058$ & $0.422$ & $0.256$ & $0.184$ \\
standard & none & \multicolumn{5}{c}{--- (no offset channel)} \\
 & indep. & \multicolumn{5}{c}{--- (no offset channel)} \\
 & corr. & \multicolumn{5}{c}{--- (no offset channel)} \\
\midrule
\multicolumn{7}{@{}l}{\emph{cheetah-run}}\\[1pt]
noncentered & none & $0.399$ & $0.396$ & $0.398$ & $0.398$ & $0.002$ \\
 & indep. & $0.137$ & $0.133$ & $0.114$ & $0.128$ & $0.013$ \\
 & corr. & $0.237$ & $0.223$ & $0.209$ & $0.223$ & $0.014$ \\
centered (ours) & none & $0.320$ & $0.322$ & $0.321$ & $0.321$ & $0.001$ \\
 & indep. & $0.341$ & $0.340$ & $0.337$ & $0.340$ & $0.002$ \\
 & corr. & $0.261$ & $0.252$ & $0.256$ & $0.256$ & $0.004$ \\
gated & none & $0.319$ & $0.320$ & $0.314$ & $0.317$ & $0.003$ \\
 & indep. & $0.340$ & $0.338$ & $0.339$ & $0.339$ & $0.001$ \\
 & corr. & $0.245$ & $0.250$ & $0.243$ & $0.246$ & $0.004$ \\
standard & none & \multicolumn{5}{c}{--- (no offset channel)} \\
 & indep. & \multicolumn{5}{c}{--- (no offset channel)} \\
 & corr. & \multicolumn{5}{c}{--- (no offset channel)} \\
\midrule
\multicolumn{7}{@{}l}{\emph{cartpole-swingup}}\\[1pt]
noncentered & none & $0.120$ & $0.120$ & $0.108$ & $0.116$ & $0.007$ \\
 & indep. & $0.008$ & $0.001$ & $0.001$ & $0.003$ & $0.004$ \\
 & corr. & $0.282$ & $0.276$ & $0.279$ & $0.279$ & $0.003$ \\
centered (ours) & none & $0.260$ & $0.261$ & $0.261$ & $0.261$ & $0.001$ \\
 & indep. & $0.248$ & $0.246$ & $0.249$ & $0.248$ & $0.001$ \\
 & corr. & $0.240$ & $0.239$ & $0.238$ & $0.239$ & $0.001$ \\
gated & none & $0.262$ & $0.260$ & $0.261$ & $0.261$ & $0.001$ \\
 & indep. & $0.246$ & $0.249$ & $0.247$ & $0.247$ & $0.001$ \\
 & corr. & $0.233$ & $0.237$ & $0.233$ & $0.234$ & $0.002$ \\
standard & none & \multicolumn{5}{c}{--- (no offset channel)} \\
 & indep. & \multicolumn{5}{c}{--- (no offset channel)} \\
 & corr. & \multicolumn{5}{c}{--- (no offset channel)} \\
\bottomrule
\end{tabular}
\caption{\textbf{Per-seed effective-offset probe $R^2$} for the four
variants run at DMC scale (standard, noncentered, centered, gated)
$\times$ 3 tasks $\times$ 3 distractor regimes ($108$ runs); the
\emph{residual} rung of the main-text ladder was not run in this matrix. The \emph{standard} predictor is monolithic and exposes no
offset channel, so it has no entry here. Seed s.d.\ is at most $0.014$
outside reacher-easy; within reacher-easy the correlated cells reach
$0.184$ (gated), $0.094$ (noncentered) and $0.067$ (centered), and the
distraction-free cells reach $0.052$ (gated) and $0.030$ (centered). The
independent-distraction column---the one the claim rests on---is tight
everywhere (s.d.\ $\le 0.013$).}
\label{tab:e3perseed}
\end{table}

\end{document}